\documentclass[twoside]{article}
\usepackage[accepted]{aistats2014}

\usepackage{balance}

\usepackage{textcase,microtype,marvosym,natbib} 
\usepackage{url}
\usepackage{tikz,pgfplots}
\pgfplotsset{compat=newest}
\pgfplotsset{plot coordinates/math parser=false}       

\usepackage[colorlinks]{hyperref}
\usepackage{amsmath,amssymb,graphicx,MnSymbol}
\usepackage{todonotes}

\definecolor{lred}{RGB}{200,0,0}
\definecolor{dred}{RGB}{130,0,0} \definecolor{dblu}{RGB}{0,0,130}
\definecolor{dgre}{RGB}{0,130,0} \definecolor{dgra}{RGB}{50,50,50}
\definecolor{mgra}{RGB}{100,100,100}
\definecolor{lgra}{RGB}{220,220,220}
\definecolor{MPG}{RGB}{000,125,122}

\usepackage{booktabs,colonequals}

\newcommand{\g}{\,|\,} 
\newcommand{\de}{\partial}
\renewcommand{\d}{\:d} 
\newcommand{\dd}{\mathrm{d}}

\newcommand{\Cov}{\operatorname{cov}} 
\newcommand{\cov}{\operatorname{cov}} 
\newcommand{\erf}{\operatorname{erf}}

\renewcommand{\Re}{\mathbb{R}}
 
\newcommand{\N}{\mathcal{N}}

\newcommand{\Trans}{^{\intercal}}

\newcommand{\argmin}{\operatorname*{arg\:min}}

\newcommand{\expmap}{\mathrm{Exp}}
\newcommand{\logmap}{\mathrm{Log}}

\newcommand{\qq}{\qquad}

\newcommand{\qqqq}{\qquad\qquad}

\renewcommand{\vec}{\boldsymbol} 
 
\usepackage[h]{esvect}
\newcommand{\vect}[1]{\vv{#1}}

\renewcommand{\O}{\mathcal{O}} \newcommand{\GP}{\mathcal{GP}}
 \newcommand{\tr}{\operatorname{tr}}

\newcommand{\f}{\vec{f}}

\usepackage{colonequals}
\newcommand{\ce}{\colonequals}
\newcommand{\ec}{\equalscolon}

\usetikzlibrary{arrows,shapes,plotmarks}

\tikzset{>=stealth'} 
\tikzstyle{graphnode} = 
   [circle,draw=black,minimum size=22pt,text centered,text
     width=22pt,inner sep=0pt] 
\tikzstyle{var}   =[graphnode,fill=white]
\tikzstyle{obs}   =[graphnode,fill=black,text=white]
\tikzstyle{fac}   =[rectangle,draw=black,fill=black!25,minimum size=5pt]
\tikzstyle{facprior} =[rectangle,draw=black,fill=black,text=white,minimum size=5pt]
\tikzstyle{edge}  =[draw=white,double=black,thick,-]
\tikzstyle{prior} =[rectangle, draw=black, fill=black, minimum size=
5pt, inner sep=0pt]
\tikzstyle{dirprior} = [circle, draw=black, fill=black, minimum
size=5pt, inner sep=0pt]

\DeclareSymbolFont{stmry}{U}{stmry}{m}{n}
\DeclareMathSymbol\leftarrowtriangle\mathrel{stmry}{"5E}
\DeclareMathSymbol\rightarrowtriangle\mathrel{stmry}{"5F}

\renewcommand{\to}{\operatorname*{\rightarrowtriangle}}

\newif\iffinal 

\iffinal
 
\else
 
\fi

\newcounter{mycomment}

\newcounter{PHcomment}

\pgfrealjobname{ProbODEs}

\begin{document}
\runningtitle{Probabilistic Solutions to Differential Equations and their Application to Riemannian Statistics}
\twocolumn[
\aistatstitle{Probabilistic Solutions to Differential Equations\\
  and their Application to Riemannian Statistics}

\aistatsauthor{Philipp~Hennig \And S{\o}ren Hauberg}
\runningauthor{Hennig \& Hauberg}
\aistatsaddress{
  Max Planck Institute for Intelligent Systems,
  Spemannstra{\ss}e 38,
  72076 T{\"u}bingen, Germany\\
  \texttt{[philipp.hennig|soren.hauberg]@tue.mpg.de}
}]

\begin{abstract}
We study a probabilistic numerical method for the solution of both
boundary and initial value problems that returns a joint Gaussian
process posterior over the solution. Such methods have concrete value
in the statistics on Riemannian manifolds, where non-analytic ordinary
differential equations are involved in virtually all computations. The
probabilistic formulation permits marginalising the uncertainty of the
numerical solution such that statistics are less sensitive to
inaccuracies. This leads to new Riemannian algorithms for mean value
computations and principal geodesic analysis. Marginalisation also
means results can be less precise than point estimates, enabling a
noticeable speed-up over the state of the art. Our approach is an
argument for a wider point that uncertainty caused by numerical
calculations should be tracked throughout the pipeline of machine
learning algorithms.
\end{abstract}

\section{Introduction}
\label{sec:introduction}

Systems of ordinary differential equations (ODEs), defined here as
equations of the form
\begin{align}
  \label{eq:1}
  c^{(n)}(t) = f(c(t),c'(t),\dots,c^{(n-1)}(t),t)
\end{align}
with $c:\Re\to\Re^D$ and $f:\Re^{nD}\times\Re\to\Re^D$ are essential
tools of numerical mathematics. In machine learning and statistics,
they play a role in the prediction of future states in dynamical
settings, but another area where they have a particularly central role
are statistical computations on Riemannian manifolds. On such
manifolds, the distance between two points is defined as the length of
the shortest path connecting them, and calculating this path requires
solving ordinary differential equations. Distances are elementary, so
virtually all computations in Riemannian statistics involve numerical
ODE solvers (Sec.~\ref{sec:riem-stat-case}). Just calculating the mean
of a dataset of $P$ points, trivial in Euclidean space, requires
optimisation of $P$ geodesics, thus the repeated solution of $P$
ODEs. Errors introduced by this heavy use of numerical machinery are
structured and biased, but hard to quantify, mostly because numerical
solvers do not return sufficiently structured error estimates.

The problem of solving ordinary differential equations is very well
studied
(\citet[][\textsection1]{hairer87:_solvin_ordin_differ_equat_i} give a
historical overview), and modern ODE solvers are highly evolved. But
at their core they are still based on the seminal work of
\citet{Runge} and \citet{Kutta}, a carefully designed process of
iterative extrapolation. In fact, ODE solvers, like other numerical
methods, can be interpreted as inference algorithms
\citep{diaconis88:_bayes,ohagan92:_some_bayes_numer_analy}, in that
they estimate an intractable quantity (the ODE's solution) given
``observations'' of an accessible quantity (values of $f$). Like all
estimators, these methods are subject to an error between estimate and
truth. Modern solvers can estimate this error
\citep[\textsection{II.3} \&
\textsection{II.4}]{hairer87:_solvin_ordin_differ_equat_i}, but
generally only as a numerical bound at the end of each local step, not
in a functional form over the space of solutions. In this paper we
study and extend an idea originally proposed by John
\citet{skilling1991bayesian} to construct
(Sec.~\ref{sec:solving-odes-with}) a class of probabilistic solvers
for both boundary value problems (BVPs) and initial value problems
(IVPs). Our version of Skilling's algorithm returns nonparametric
posterior probability measures over the space of solutions of the ODE.
This kind of structured uncertainty improves Riemannian statistics,
where the solution of the ODE is a latent ``nuisance'' variable. As
elsewhere, it is beneficial to replace point estimates over such
variables with probability distributions and marginalise over them
(Sec.~\ref{sec:riem_prob}).  A second, remarkable result is that,
although the probabilistic solver presented here provides additional
functionality, when applied to concrete problems in Riemannian
statistics, it can actually be faster than standard boundary value
problem solvers (as implemented in Matlab). One reason is that the
answers returned by the probabilistic solver can be less precise,
because small errors are smoothed out when marginalised over the
posterior uncertainty. As the cost of Riemannian statistics is
dominated by that of solving ODEs, time saved by being ``explicitly
sloppy'' in the probabilistic sense is of great value.
Sec.~\ref{sec:experiments} offers some empirical evaluations. A Matlab
implementation of the solver is published alongside this
manuscript\footnote{\url{http://probabilistic-numerics.org/ODEs.html}}.

The focus of this text is on the concrete case of Riemannian
statistics and ODEs. But we believe it to be a good example for
the value of probabilistic numerical methods to machine learning and
statistics more generally. Treating numerical algorithms as black
boxes ignores their role as error sources and can cause unnecessary
computational complexity. Our results show that algorithms designed
with probabilistic calculations in mind can offer qualitative and
quantitative benefits. In particular, they can propagate uncertainty
through a numerical computation, providing a joint approximate
distribution between input and output of the computation.


\section{Riemannian Statistics}
\label{sec:riem-stat-case}

Statistics on Riemannian manifolds is becoming increasingly acknowledged
as it allows for statistics in constrained spaces, and in spaces
with a smoothly changing metric. Examples include analysis of shape
\citep{Freifeld:ECCV:2012, fletcher2004principal}, human poses
\citep{hauberg:imavis:2011} and image features
\citep{Porikli:CVPR:2006}. Here we focus on Riemannian statistics in
\emph{metric learning}, where a smoothly changing metric 
locally adapts to the data \citep{hauberg:nips:2012}.

Let $M(x)\in\Re^{D\times D}$ denote a positive definite, smoothly
changing, metric tensor function at $x\in\Re^D$. It locally defines a
norm $\|y\|^2 _{M(x)} = y\Trans M(x)y$. This norm in turn defines
\citep{berger07} the \emph{length} of a curve $c:[0,1]\to\Re^D$ as
$\mathrm{Length}(c) := \int_0 ^1 \|c'(t)\|_{M(c(t))} \dd t$. For two
points $a$, $b$, the shortest connecting curve, $c = \argmin_{\hat{c}}
\mathrm{Length} (\hat{c})$ is known as a \emph{geodesic}, and defines
the \emph{distance} between $a$ and $b$ on the manifold. Using the
Euler-Lagrange equation, it can be shown \citep{hauberg:nips:2012} that
geodesics must satisfy the system of homogenous second-order boundary
value problems
\begin{align}
  \notag c''(t) &= -\frac{1}{2}M^{-1}(c(t)) \left[ \frac{\de
      \vect{M}(c(t))}{\de c(t)}\right]\Trans [c'(t)\otimes c'(t)]\\
  \label{eq:2}
    &=: f(c(t),c'(t)),\text{ with } c(0)=a, c(1)=b.
\end{align}
where the notation $\vv{M}$ indicates the vectorization operation
returning a vector $\in\Re^{D^2}$, formed by stacking the elements of
the $D\times D$ matrix $M$ row by row, and $\otimes$ is the Kronecker
product, which is associated with the vectorization operation
\citep{loan00:_kronec}. The short-hand $f(c(t),c'(t))$ is used
throughout the paper. For simple manifolds this BVP can be solved
analytically, but most models of interest require numerical solutions.

The geodesic starting at $a$ with initial direction $v$ is given by
the initial value problem
\begin{equation}
  \label{eq:geo_ivp}
  c''(t) = f(c(t),c'(t)), \quad c(0) = a \quad\mathrm{and}\quad c'(0) = v.
\end{equation}
By the Picard--Lindel{\"o}f theorem \citep{tenenbaum}, this curve
exists and is locally unique. As $v$ is the derivative of a curve on
the manifold at $a$, it belongs to the \emph{tangent space} of the
manifold at $a$. Eq.~\eqref{eq:geo_ivp} implies that geodesics can be
fully described by their tangent $v$ at $c(0)$. This observation is a
corner-stone of statistical operations on Riemannian manifolds, which
typically start with a point $\mu$ (often the mean), compute geodesics
from there to all data points, and then perform \emph{Euclidean}
statistics on the tangents $v_{1:P}$ at $\mu$. This implicitly gives
statistics of the geodesics.

These concepts are formalised by the \emph{exponential} and
\emph{logarithm} maps, connecting the tangent space and the
manifold. The exponential map $\expmap_a (v)$ returns the end-point of
the geodesic $c$ satisfying Eq.~\eqref{eq:geo_ivp} such that
$\mathrm{Length}(c) = \| v \|$.  The logarithm map $\logmap_a(b)$ is
the inverse of the exponential map. It returns the initial direction
of the geodesic from $a$ to $b$, normalised to the length of the
geodesic, i.e.
\begin{equation}\label{eq:logmap}%
  \logmap_a(b) = \frac{c'(0)}{\| c'(0) \|} \mathrm{Length}(c) .
\end{equation}
The following paragraphs give examples for how this theory is used in
statistics, and how the use of numerical solvers in these problems
affect the results and their utility, thus motivating the use of a
probabilistic solver.

\paragraph{Means on Manifolds}\label{sec:means}
The most basic statistic of interest is the mean, which is defined as
the minimiser of the sum of squared distances on the manifold. A
global minimiser is called the \emph{Frech\'et mean}, while a local
minimiser is called a \emph{Karcher mean} \citep{Pennec99}.  In
general, closed-form solutions do not exist and gradient descent is
applied; the update rule of $\mu_k$ with learning rate $\alpha$ is
\citep{Pennec99}
\begin{equation}
  \label{eq:mean_update}
  \mu_{k+1} = \expmap_{\mu_k} \left( \alpha \frac{1}{P}\sum_{i=1}^P \logmap_{\mu_k} (x_i) \right).
\end{equation}
One practical issue is that, in each iteration, the exponential and
logarithm maps need to be computed numerically. Each evaluation of
Eq.~\eqref{eq:mean_update} requires the solution of $P$ BVPs and one
IVP. Practitioners thus perform only a few iterations rather than run
gradient descent to convergence. Intuitively, it is clear that these
initial steps of gradient descent do not require high numerical
precision. The solver should only control their error sufficiently to
find a rough direction of descent.

\paragraph{Principal Geodesic Analysis}\label{sec:pga}
A generalisation of principal component analysis (PCA) to Riemannian
manifolds is \emph{principal geodesic analysis (PGA)}
\citep{fletcher2004principal}. It performs ordinary PCA in the tangent
space at the mean, i.e.\ the \emph{principal directions} are found as
the eigenvectors of $\logmap_{\mu} (x_{1:P})$, where $\mu$ is the mean
as computed by Eq.~\eqref{eq:mean_update}. The principal direction $v$
is then mapped back to the manifold to form \emph{principal geodesics}
via the exponential map, i.e.\ $\gamma (t) = \expmap_{\mu} (t v)$.

Again, this involves solving for $P$ geodesics: one from each $x_i$ to
the mean. As the statistics are performed on the (notoriously
unstable) derivatives of the numerical solutions, here, high-accuracy
solvers are required. These computational demands only allow for
low-sample applications of PGA.

\paragraph{Applications in Metric Learning}\label{sec:appl-metr-learn}
The original appeal of Riemannian statistics is that many types of
data naturally live on known Riemannian manifolds
\citep{Freifeld:ECCV:2012, fletcher2004principal,Porikli:CVPR:2006},
for example certain Lie groups. But a recent generalisation is to
estimate a smoothly varying metric tensor from the data, using metric
learning, and perform Riemannian statistics on the manifold implied by
the smooth metric \citep{hauberg:nips:2012}. The idea is to regress a
smoothly changing metric tensor from learned metrics $M_{1:R}$ as
\begin{align} M(x) &= \sum_{r=1}^R
  \frac{\tilde{w}_r(x)}{\sum_{j=1}^R \tilde{w}_j(x)} M_r,
  \qquad\mathrm{where}\\
  \tilde{w}_r(x) &= \exp \left(
  -\frac{\rho}{2} (x - \mu_r)\Trans M_r (x - \mu_r) \right).
  \label{eq:metric}
\end{align}
As above, geodesics on this manifold require numeric solutions; hence
this approach is currently only applicable to fairly small
problems. The point estimates returned by numerical solvers, and their
difficult to control computational cost, are both a conceptual and
computational roadblock for the practical use of Riemannian
statistics. A method returning explicit probabilistic error estimates,
at low cost, is needed.


\section{Solving ODEs with Gaussian process priors}
\label{sec:solving-odes-with}

Writing in 1991, John Skilling proposed treating initial value
problems as a recursive Gaussian regression problem: Given a Gaussian
estimate for the solution of the ODE, one can evaluate a linearization
of $f$, thus construct a Gaussian uncertain evaluation of the
derivative of $c$, and use it to update the belief on $c$. The
derivations below start with a compact introduction to this idea in a
contemporary notation using Gaussian process priors, which is
sufficiently general to be applicable to both boundary and initial
values. Extending Skilling's method, we introduce a way to iteratively
refine the estimates of the method similar to implicit Runge-Kutta
methods, and a method to infer the hyperparameters of the
algorithm.

The upshot of these derivations is a method returning a nonparametric
posterior over the solution of both boundary and initial value
problems. Besides this more expressive posterior uncertainty, another
functional improvement over contemporary solvers is that these kind of
algorithms allow for Gaussian uncertainty on the boundary or initial
values, an essential building block of probabilistic computations.

In parallel work, \citet{2013arXiv1306.2365C} have studied
several theoretical aspects of Skilling's method thoroughly. Among
other things, they provide a proof of consistency of the probabilistic
estimator and argue for the uniqueness of Gaussian priors on ODE and
PDE solutions. Here, we focus on the use of such probabilistic
numerical methods within a larger (statistical) computational
pipeline, to demonstrate the strengths of this approach, argue for its
use, and highlight open research questions. We also offer a minor
theoretical observation in Section \ref{sec:digr-relat-runge}, to
highlight a connection between Skilling's approach and the classic
Runge-Kutta framework that has previously been overlooked.

Conceptually, the algorithm derived in the following applies to the
general case of Eq.~(\ref{eq:1}). For concreteness and simplicity of
notation, though, we focus on the case of a second-order ODE. All
experiments will be performed in the Riemannian context, where the
object to be inferred is the geodesic curve $c:t\in\Re\mapsto
c(t)\in\Re^D$ consistent with Eqs.~(\ref{eq:2}) or (\ref{eq:geo_ivp}):
Distance calculations between two points are boundary value problems
of the form $c(0) = a$, $c(1) = b$. Shifts of variables along
geodesics are initial value problems of the form $c(0) = a$, $c'(0) =
v$ where $v$ is the length of the geodesic. The probabilistically
uncertain versions of these are
\begin{align}
  \label{eq:3}
  p(c(0),c(1)) &= \N(c(0); a,\Delta_a)\N(c(1); b,\Delta_b), 
  \text{ or}\\
  \label{eq:19}
  p(c(0),c'(0)) &= \N(c(0); a,\Delta_a)\N(c'(0); v,\Delta_v).
\end{align}
with positive semidefinite covariance matrices
$\Delta_a,\Delta_b,\Delta_v\in\Re^{D\times D}$.

\subsection{Prior}
\label{sec:prior}

We assign a joint Gaussian process prior over the elements of the
geodesic, with a covariance function factorizing between its input and
output dependence ($c_i(t)$ is the $i$-th element of $c(t)\in\Re^D$):
\begin{gather}
  \label{eq:4}
  p(c(t)) = \GP(c;\mu_c,V\otimes k), \qq\text{that is,}\\
  \Cov(c_i(t_1),c_j(t_2)) = V_{ij} k(t_1,t_2),\notag
\end{gather}
with a positive semidefinite matrix $V$ -- the covariance between
output dimensions, and a positive semidefinite kernel
$k:\Re\times\Re\to\Re$ defining the covariance between scalar
locations along the curve. The geodesic problem only involves values
$t\in[0,1]$, but this restriction is not necessary in general. The
mean functions $\mu_c:\Re\to\Re^D$ have only minor effects on the
algorithm. For BVPs and IVPs, we choose linear functions $\mu(t) = a +
(b-a)t$ and $\mu(t) = a + vt$, respectively. There is a wide space of
covariance functions available for Gaussian process regression
\citep[][\textsection 4]{RasmussenWilliams}. In principle, every
smooth covariance function can be used for the following derivations
(more precisely, see Eq.~(\ref{eq:5}) below). For our implementation
we use the popular square-exponential (aka.~radial basis function,
Gaussian) kernel $k(t_1,t_2) =
\exp\left[-(t_1-t_2)^2/(2\lambda^2)\right]$. This amounts to the prior
assumption that geodesics are smooth curves varying on a length scale
of $\lambda$ along $t$, with output covariance
$V$. Sec.~\ref{sec:choice-hyperp} below explains how to choose these
hyperparameters. 

Gaussian processes are closed under linear transformations. For any
linear operator $A$, if $p(x)=\GP(x;\mu,k)$ then $p(Ax) =
\GP(Ax;A\mu,AkA\Trans)$. In particular, a Gaussian process prior on
the geodesics $c$ thus also implies a Gaussian process belief over any
derivative $\de^n c(t)/\de t^n$ (see
\citet[][\textsection9.4]{RasmussenWilliams}), with mean function
$\de^n \mu(t)/\de t^n$ and covariance function
\begin{equation}
  \label{eq:5}
  \cov\left(\frac{\de^n c_i(t_1)}{\de t_1 ^n},\frac{\de^n c_j(t_2)}{\de
      t^n _2}\right) = V_{ij}\frac{\de^{2n} k(t_1,t_2)}{\de t_1^{n}
    \de t_2 ^n},
\end{equation}
provided the necessary derivatives of $k$ exist, as they do for the
square-exponential kernel $\forall n\in\mathbb{N}$. Explicit algebraic
forms for the square-exponential kernel's case can be found in the
supplements.

The boundary or initial values $a,b$ or $a,v$, can be incorporated
into the Gaussian process belief over $c$ analytically via the factors
defined in Eqs.~(\ref{eq:3}) and (\ref{eq:19}). For the case of
boundary value problems, this gives a Gaussian process conditional
over $c$ with mean and covariance functions (c.f.~supplements)
\begin{align}
  \label{eq:7}
  \mu_{c\g a,b}(t) &= \mu_c(t) + 
  \left[V\otimes
  \begin{pmatrix}
    k_{t0} & k_{t1}
  \end{pmatrix}\right]\\
\notag
\cdot&\underbrace{\left[V\otimes
\begin{pmatrix}
  k_{00} & k_{01} \\ k_{10} & k_{11}
\end{pmatrix} +
\begin{pmatrix}
  \Delta_a & 0 \\ 0 & \Delta_b
\end{pmatrix}
\right] ^{-1}}_{=:G^{-1}}
  \begin{pmatrix}
    a-\mu_c(0)\\b - \mu_c(1)
  \end{pmatrix}
\end{align}
\begin{multline}
\label{eq:6}
\text{and } \cov(c(t_1),c(t_2)\g a,b) = V\otimes k_{t_1,t_2} \\ \notag
  - \left[V\otimes
  \begin{pmatrix}
    k_{t_1 0} & k_{t_1  1}
  \end{pmatrix}\right]
G^{-1} \left[V\otimes
  \begin{pmatrix}
    k_{0t_2} \\ k_{1t_2}
  \end{pmatrix}\right].
\end{multline}
For IVPs, $k_{t_1 1},k_{1t_2}$ have to be replaced with
$k'_{t_10},k'_{0t_2}$, $b$ with $v$ and $\mu_c(1)$ with $\mu'_c (0)$.

\subsection{Likelihood}
\label{sec:likelihood}

At this point we pick up Skilling's idea: Select a sequence
$t_1,\dots,t_N\in [0,1]$ of representer points. The algorithm now
moves iteratively from $t_1$ to $t_N$. At point $t_i$, construct
estimates $[\hat{c}_i,\hat{c}'_i$] for the solution of the ODE and its
derivative. The mean (and simultaneously the most probable) estimate
under the Gaussian process belief on $c$ is the mean function, so we
choose $\hat{c}_i = \mu_{c\g a,b}(t_i)$, $\hat{c}' _i = \de_t\mu_{c\g
  a,b}(t_i)$. This estimate can be used to construct an
``observation'' $y_i := f(\hat{c}_i,\hat{c}'_i) \approx c''(t_i)$ of
the curvature of $c$ along the solution. For our application we use
the $f$ defined in Eq.~(\ref{eq:2}). This is homogeneous, so $t_i$
does not feature as a direct input to $f$, but other $f$'s, including
inhomogeneous ones, can be treated in the exact same way. The idea of
constructing such approximately correct evaluations of $f$ along the
solution is also at the heart of classic ODE solvers such as the
Runge-Kutta family (see Section \ref{sec:digr-relat-runge}).
Skilling's idea, however, is to probabilistically describe the error
of this estimation by a Gaussian distribution: Assume that we have
access to upper bounds $U,U'\in\Re^D$ on the absolute values of the
elements of the Jacobians of $f$ with respect to $c$ and $c'$:
\begin{equation}
  \label{eq:9}
  \left|\frac{\de f_j(c(t),c'(t))}{\de c_i(t)}\right| < U_{ij}
  \quad\text{and}\quad\left|\frac{\de f_j(c(t),c'(t))}{\de c' _i(t)}\right| <
  U' _{ij}
\end{equation}
for all $t\in[0,1]; i,j\in(1,\dots,D)$.  In the specific case of the
Riemannian problem on learned metrics, these bounds arise from
analysis of Eq.~(\ref{eq:metric}). The uncertainty on $c,c'$ around
$\hat{c}_i$, $\hat{c}_i '$ is Gaussian, with covariance (see also
supplements)
\begin{align}
  \label{eq:11}&
  \begin{pmatrix}
    \Sigma^{cc} _i & \Sigma^{cc'} _i \\ \Sigma^{c'c} _i &
    \Sigma^{c'c'} _i
  \end{pmatrix}
:= V\otimes
  \begin{pmatrix}
    k_{t_it_i} & k'_{t_i,t_i} \\ k'_{t_it_i} & k''_{t_it_i}
  \end{pmatrix}\\ \notag
  &- \left[V\otimes
  \begin{pmatrix}
    k_{t_i 0} & k_{t_i  1}\\
    k'_{t_i 0} & k'_{t_i  1}\\
  \end{pmatrix}\right]
G^{-1} \left[V\otimes
  \begin{pmatrix}
    k_{0t_i} & k'_{0t_i}\\ k_{1t_i} & k'_{1t_i}
  \end{pmatrix}\right] \in\Re^{2D \times 2D}.
\end{align}
Considering this prior uncertainty on $c,c'$ to first order, the error
caused by the estimation of $c,c'$ used to construct the observation
$y_i$ is bounded by a zero-mean Gaussian, with covariance (in $\Re^{D\times D}$)
\begin{equation}
  \label{eq:10}
  \Lambda_i = U\Trans\Sigma_i ^{cc} U + \left|{U'}\Trans\Sigma_i ^{c'c} U\right| +
  \left|U\Trans\Sigma_i ^{cc'} U'\right| + {U'}\Trans\Sigma_i ^{c'c'} U'.
\end{equation}
This uncertain observation $y_i$ of $c''(t_i)$ is included into the
belief over $c$ as the likelihood factor
\begin{equation}
  \label{eq:12}
  p(y_i\g c) = \N(y_i;c''(t_i),\Lambda_i).
\end{equation}
The algorithm now moves to $t_{i+1}$ and repeats.

\begin{figure*}
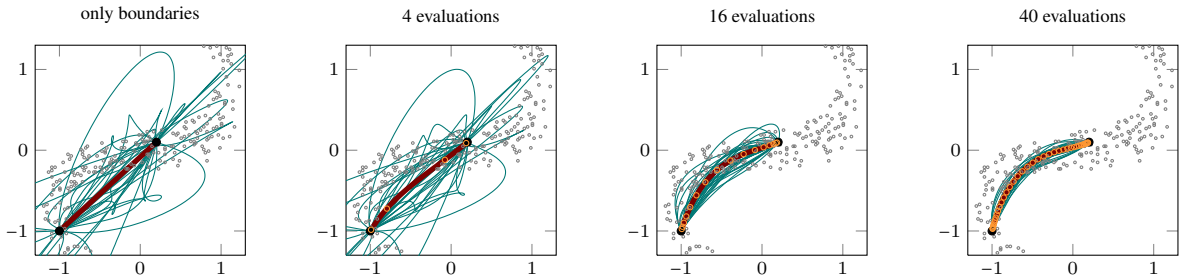

  \centering
  \scriptsize\hfill%
   \beginpgfgraphicnamed{figures/exp011_step_00-external}%
   \input{figures/exp011_step_00.tikz}%
   \endpgfgraphicnamed%
 \hfill
   \beginpgfgraphicnamed{figures/exp011_step_04-external}%
   \input{figures/exp011_step_04.tikz}%
   \endpgfgraphicnamed%
 \hfill
   \beginpgfgraphicnamed{figures/exp011_step_16-external}%
   \input{figures/exp011_step_16.tikz}%
   \endpgfgraphicnamed%
 \hfill
   \beginpgfgraphicnamed{figures/exp011_step_40-external}%
   \input{figures/exp011_step_40.tikz}%
   \endpgfgraphicnamed%
 \hfill
  \vspace{-3mm}
  \caption{Convergence of probabilistic ODE solver as number of
    evaluation points increases. Each panel shows a dataset (grey
    points) whose distribution determines a Riemannian manifold. The
    algorithm infers the geodesic connecting two points (black) on
    this manifold. Mean of inferred Gaussian process posterior in
    red. Evaluation points of the ODE in orange. Green lines are
    samples drawn from the posterior belief over geodesics. Left-most
    panel: no ODE evaluations (only boundary values).  The supplements
    contain an animated version of this figure.}
  \label{fig:exp011}
\end{figure*}

\subsection{Posterior}
\label{sec:posterior}

After $N$ steps, the algorithm has collected $N$ evaluations
$y_i=f(\hat{c}_i,\hat{c}'_i)$, and constructed equally many
uncertainty estimates $\Lambda_i$. The marginal distribution over the
$y_i$ under the prior (\ref{eq:4}) is Gaussian with mean $[\de^2
\mu_c(t_1)/\de t^2,\dots,\de^2 \mu_c(t_N)/\de t^2]\Trans$ and
covariance
\begin{align}
  \label{eq:13}
  \Gamma &:= V \otimes k^{(4)} _{TT} + \Lambda\\
  \notag
  \text{where}\quad k^{(4)}_{TT}&\in\Re^{N\times N}
\quad\text{with}\quad k^{(4)} _{TT,ij} = \frac{\de^4
  k(t_i,t_j)}{\de t_i^2 \de t_j ^2}.
\end{align}
The matrix $\Lambda\in\Re^{DN\times DN}$ is a rearranged form of the
block diagonal matrix whose $N$-th block diagonal element is the
matrix $\Lambda_i\in\Re^{D\times D}$. The posterior distribution over
$c$, and the output of the algorithm, is thus a Gaussian process with
mean function
\begin{multline}
  \label{eq:14}
  \mu_{c\g a,b,Y}(t) = \mu_c(t) +
  \left[V\otimes
  \begin{pmatrix}
    k_{t0} & k_{t1} & k''_{tT}
  \end{pmatrix}\right]\cdot\\
\underbrace{
\left[V\otimes
\begin{pmatrix}
  k_{00} & k_{01} & k'' _{0T} \\ k_{10} & k_{11} & k'' _{1T}\\
  k''_{T0} & k''_{T1} & k^{(4)} _{TT}
\end{pmatrix} +
\begin{pmatrix}
  \Delta_a & 0 & 0\\ 0 & \Delta_b & 0 \\ 0 & 0 & \Lambda
\end{pmatrix}
\right]^{-1}}_{=:\Gamma^{-1}} \\
\cdot
  \begin{bmatrix}
    a-\mu_c(0) & b - \mu_c(1) & Y - \mu'' _c(T)
  \end{bmatrix}\Trans
\end{multline}
(where $k'' _{0 t_j} = \de k(0,t_j)/\de t_j^2 $, etc.) and covariance
function (see also supplements)
\begin{multline}
\label{eq:15}
\cov(c(t_1),c(t_2)\g a,b,Y) = V\otimes k_{t_1,t_2}\\
- \left[V\otimes
  \begin{pmatrix}
    k_{t0} & k_{t1} & k''_{tT}
  \end{pmatrix}\right]
\Gamma^{-1} \left[V\otimes
  \begin{pmatrix}
    k_{0t_2} \\ k_{1t_2} \\ k'' _{T t_2}
  \end{pmatrix}\right].
\end{multline}
Fig.~\ref{fig:exp011} shows conceptual sketches. Crucially, this
posterior is a joint distribution over the entire curve, not just
about its value at a particular value along the curve. This sets it
apart from the error bounds returned by classic ODE solvers, and means
it can be used directly for Riemannian statistics
(Sec.~\ref{sec:riem_prob}).

\subsubsection{Iterative Refinements}
\label{sec:iter-refin}

We have found empirically that the estimates of the method can be
meaningfully (albeit not drastically) improved by a small number of
iterations over the estimates $\hat{c}_i,\hat{c}'_i$ after the first
round, without changing the Gram matrix $\Gamma$. In other words, we
update the mean estimates used to construct $\hat{c},\hat{c}'$ based
on the posterior arising from later evaluations. This idea is related
to the idea behind implicit Runge-Kutta methods \citep[][\textsection
II.7]{hairer87:_solvin_ordin_differ_equat_i}. It is important not to
update $\Gamma$ based on the posterior uncertainty in this process;
this would lead to strong over-fitting, because it ignores the
recursive role earlier evaluation play in the construction of later
estimates.

\subsection{Choice of Hyperparameters and Evaluation Points}
\label{sec:choice-hyperp}

The prior definitions of Eq.~(\ref{eq:4}) rely on two hyperparameters
$V$ and $\lambda$. The output covariance $V$ defines the overall scale
on which values $c(t)$ are expected to vary. For curves within a
dataset $x_{1:P} \in \Re^D$, we assume that the curves are typically
limited to the Euclidean domain covered by the data, and that
geodesics connecting points close within the dataset typically have
smaller absolute values than those geodesics connecting points far
apart within the dataset. For curves between $a,b$, we thus set $V$ in
an empirical Bayesian fashion to $V = \left[(a-b)\Trans S_x(a-b)
\right]\cdot S_x \in\Re^{D\times D}$, where $S_x$ is the sample
covariance of the dataset $x_{1:P}$. For IVPs, we replace $(a-b)$ in
this expression by $v$.

The width $\lambda$ of the covariance function controls the regularity
of the inferred curve. Values $\lambda\ll 1$ give rise to rough
estimates associated with high posterior uncertainty, while $\lambda
\gg 1$ gives regular, almost-straight curves with relatively low
posterior uncertainty. Following a widely adapted approach in Gaussian
regression \citep[][\textsection{5.2}]{RasmussenWilliams}, we choose
$\lambda$ to maximise its marginal likelihood under the observations
$Y$. This marginal is Gaussian. Using the Gram matrix $\Gamma$ from
Eq.~(\ref{eq:14}), its logarithm is
\begin{multline}
  \label{eq:18}
  -2\log p(Y,a,b\g \lambda) =
      \begin{pmatrix}
    a-\mu_c(0)\\b - \mu_c(1)\\
    Y - \mu'' _c(T)
  \end{pmatrix}\Trans \Gamma^{-1}   \begin{pmatrix}
    a-\mu_c(0)\\b - \mu_c(1)\\
    Y - \mu'' _c(T)
  \end{pmatrix}\\
  + \log|\Gamma| + (2+N)\log 2\pi.
\end{multline}
A final design choice is the location of the evaluation points
$t_i$. The optimal choice of the grid points depends on $f$ and is
computationally prohibitive, so we use a pre-determined grid.  For
BVP's, we use sigmoid grids such as $t_i = 0.5 (1 + \erf(-1.5 + 3i/N))$
as they emphasise boundary areas; for IVP's we use linear grids.
But alternative choices give acceptably good results as well.

\begin{figure}[h]
  \scriptsize\centering
   \beginpgfgraphicnamed{figures/exp008_fig1-external}%
   \input{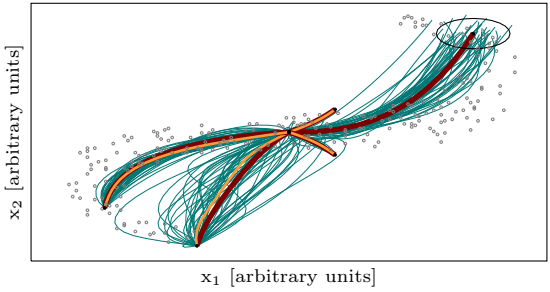}%
   \endpgfgraphicnamed%
 
  \vspace{-4mm}
  \caption{Solving five BVPs within a dataset (grey points). Orange:
    solutions found by Matlab's \texttt{bvp5c}. For the probabilistic solver,
    30 posterior samples (thin green), and mean as thick red
    line. Simpler short curves lead to confident predictions, while
    the challenging curve to the bottom left has drastic
    uncertainty. Top right curve demonstrates the ability, not
    offered by classic methods, to deal with uncertain boundary
    values 
    (the black ellipse represents two standard deviations).}
\label{fig:exp008}
\end{figure}

\paragraph{Cost}
\label{sec:cost}
Because of the inversion of the $((N+2)D)\times ((N+2)D)$ matrix
$\Gamma$ in Eq.~(\ref{eq:14}), the method described here has
computational cost $\O(D^3N^3)$. This may sound costly. However, both
$N$ and $D$ are usually small numbers for problems in Riemannian
statistics. Modern ODE solvers rely on a number of additional
algorithmic steps to improve performance, which lead to considerable
computational overhead. Since point estimates require high precision,
the computational cost of existing methods is quite high in
practice. We will show below that, since probabilistic estimates can
be marginalised within statistical calculations, our algorithm in fact
improves computation time.

\subsection{Relationship to Runge-Kutta methods}
\label{sec:digr-relat-runge}

The idea of iteratively constructing estimates for the solution and
using them to extrapolate may seem ad hoc. In fact, the step from the
established Runge-Kutta (RK) methods to this probabilistic approach is
shorter than one might think. RK methods proceed in extrapolation
steps of length $h$. Each such step consists of $N$ evaluations $y_i
=f(\hat{c}_i,t_i)$. The estimate $\hat{c}_i$ is constructed by a
\emph{linear} combination of the previous evaluations and the initial
values: $\hat{c}_i = w_0c_0 + \sum_{j=1} ^{i-1} w_j y_j $. This is
exactly the structure of the mean prediction of Eq.~(\ref{eq:14}). The
difference lies in the way the weights $w_j$ are constructed. For RK,
they are constructed carefully such that the error between the true
solution of the ODE and the estimated solution is
$\O(h^{-N})$. Ensuring this is highly nontrivial, and often cited as a
historical achievement of numerical mathematics
\citep[][\textsection{II.5}]{hairer87:_solvin_ordin_differ_equat_i}.
In the Gaussian process framework, the $w_i$ arise much more
straightforwardly, from Eq.~(\ref{eq:14}), the prior choice of kernel,
and a (currently ad hoc) exploration strategy in $t$. The strength of
this is that it naturally gives a probabilistic generative model for
solutions, and thus various additional functionality. The downside is
that, in this simplistic form, the probabilistic solver can not be
expected to give the same local error bounds as RK methods. In fact
one can readily construct examples where our specific form of the
algorithm has low order. It remains a challenging research question
whether the frameworks of nonparametric regression and Runge-Kutta
methods can be unified under a consistent probabilistic concept.


\section{Probabilistic Calculations on Riemannian Manifolds}\label{sec:riem_prob}
We briefly describe how the results from Sec.~\ref{sec:solving-odes-with} 
can be combined with the concepts from
Sec.~\ref{sec:riem-stat-case} to define tools for Riemannian
statistics that marginalise numerical uncertainty of the solver.

\paragraph{Exponential maps} Given an initial point $a \sim
\N (\bar{a}, \Sigma_a)$ and initial direction $v \sim \N(\bar{v},
\Sigma_v)$, the exponential map can be computed by solving an IVP
(\ref{eq:geo_ivp}) as described in
Sec.~\ref{sec:solving-odes-with}. This makes direct use of the ability
of the solver to deal with uncertain boundary and initial values.

\vspace{-3mm}
\paragraph{Logarithm maps} The logarithm map \eqref{eq:logmap} is
slightly more complicated to compute as the length of a geodesic
cannot be computed linearly. We can, however, sample from the
logarithm map as follows. Let $a \sim \N (\bar{a}, \Sigma_a)$ and $b
\sim \N (\bar{b}, \Sigma_b)$ and construct the geodesic Gaussian
process posterior. Sample $c_s$ and $c_s'$ from this process and
compute the corresponding logarithm map using
Eq.~\eqref{eq:logmap}. The length of $c_s$ is evaluated using standard
quadrature.  The mean and covariance of the logarithm map can then be
estimated from the samples. Resorting to a sampling method might seem
expensive, but in practice the runtime is dominated by the BVP solver.

\vspace{-3mm}
\paragraph{Computing Means} 
The mean of $x_{1:P}$ is computed using gradient descent
\eqref{eq:mean_update}, so we need to track the uncertainty
through this optimisation. Let $\mu_k \sim \N (\bar{\mu}_k,
\Sigma_{\mu_i})$ denote the estimate of the mean at iteration $i$ of
the gradient descent scheme. First, compute the mean and covariance of
\begin{align}
  v = \frac{1}{P} \sum_i \logmap_{\mu_k} (x_i)
\end{align}
by sampling as
above. Assuming $v$ follows a Gaussian distribution with the estimated
mean and covariance, we then directly compute the distribution of
\begin{align}
  \mu_{k+1} = \expmap_{\mu_k} (\alpha v).
\end{align}
Probabilistic optimisation
algorithms \citep{StochasticNewton}, which can directly return
probabilistic estimates for the shape of the gradient
(\ref{eq:mean_update}), are a potential alternative.

\vspace{-3mm}
\paragraph{Computing Principal Geodesics} 
Given a mean $\mu \sim \N (\bar{\mu}, \Sigma_{\mu})$, estimate the mean and
covariance of $\logmap_{\mu} (x_{1:P})$ and use these to estimate the mean and
covariance of the principal direction $v$. The principal geodesic can
then be computed as $\gamma = \expmap_{\mu} (v)$.


\section{Experiments}
\label{sec:experiments}

\begin{figure*}
  \centering \scriptsize
  \mbox{
  \hspace{-5mm}
   \beginpgfgraphicnamed{figures/exp015_fig1b-external}%
   \input{figures/exp015_fig1b.tikz}%
   \endpgfgraphicnamed%
 \hspace{9mm}
   \beginpgfgraphicnamed{figures/exp015_fig1c-external}%
   \input{figures/exp015_fig1c.tikz}%
   \endpgfgraphicnamed%
   \beginpgfgraphicnamed{figures/exp015_fig1d-external}%
   \input{figures/exp015_fig1d.tikz}%
   \endpgfgraphicnamed%
 
  }
  \vspace{-6mm}
  \caption{Results on 1000 images of the digit 1 from the MNIST
    handwritten digits data set, projected into its first two
    principal components. Left: Several geodesics from the mean
    computed using the probabilistic solver (red mean predictions,
    green samples) as well as Matlab's \texttt{bvp5c}
    (orange). Middle: The principal geodesics estimated using the same
    two solvers. Right: Running time of a mean value computation using
    different solvers as a function of the dimensionality.}
  \label{fig:mnist}
\end{figure*}

This section reports on experiments with the suggested
algorithms. It begins with an illustrative experiment on the
well-known MNIST handwritten data set\footnote{\tt
  http://yann.lecun.com/exdb/mnist/}, followed by a study of the
running time of the algorithms on the same data. Finally, we use the
algorithms for visualising changes in human body shape.

\subsection{Handwritten Digits}\label{sec:mnist}
Fig.~\ref{fig:mnist} (next page, for layout reasons), left and centre,
show 1000 images of the digit 1 expressed in the first two principal
components of the data. This data roughly follows a one-dimensional
nonlinear structure, which we estimate using PGA. To find geodesics
that follow the local structure of the data, we learn local metric
tensors as local inverse covariance matrices, and combine them to form
a smoothly changing metric using Eq.~\ref{eq:metric}.  The local
metric tensors are learned using a standard EM algorithm for Gaussian
mixtures \citep{mclachlan97:_em_algor_exten}.

\begin{figure}
  \centering \scriptsize
  \mbox{
   \beginpgfgraphicnamed{figures/exp018_fig1-external}%
   \input{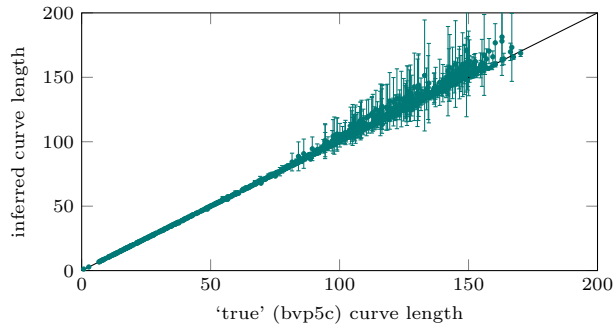}%
   \endpgfgraphicnamed%
  }
  \vspace{-4mm}
  \caption{Inferred curve lengths for the 1000 geodesics from
    Figure \ref{fig:mnist}. The probabilistic estimates for the
    lengths (mean $\pm$ 2 standard deviations) are plotted against the
    (assumed reliable) point estimates of Matlab's {\tt bvp5c}.}
  \label{fig:MNISTquant:left}
\end{figure}

For comparison, we consider a state-of-the-art BVP solver as
implemented in Matlab in \texttt{bvp5c} (a four-stage implicit
Runge-Kutta method \citep[][\textsection
II.7]{hairer87:_solvin_ordin_differ_equat_i}). This does not provide
an uncertain output, so we resort to the deterministic Riemannian
statistics algorithms (Sec.~\ref{sec:riem-stat-case}).
Fig.~\ref{fig:mnist}, left, shows a few geodesics from the mean
computed by our solver as well as \texttt{bvp5c}. The geodesics
provided by the two solvers are comparable and tend to follow the
structure of the data. Fig.~\ref{fig:mnist}, centre, shows the
principal geodesics of the data computed using the different solvers;
the results are comparable. This impression is supported by a more
close comparison of the results: Figure \ref{fig:MNISTquant:left}
shows curve lengths achieved by the two solvers. While \texttt{bvp5c}
tends to perform slightly better, in particular for long curves, the
error estimates of the probabilistic solvers reflect the algorithm's
imprecision quite well. 

\begin{figure}
  \centering \scriptsize
  \mbox{
   \beginpgfgraphicnamed{figures/exp018_fig2-external}%
   \input{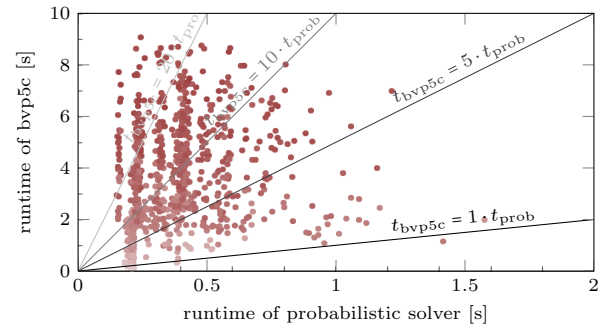}%
   \endpgfgraphicnamed%
 }
  \vspace{-4mm}
  \caption{Runtime of {\tt bvp5c} against that of the probabilistic solver
    for the same geodesics (note differing abscissa and
    ordinate scales). The color of points encodes the length of the
    corresponding curve, from 0 (white) to 200 (dark red), for
    comparison with Fig.~\ref{fig:MNISTquant:left}.}
  \label{fig:MNISTquant:right}
\end{figure}

Even within this simple two-dimensional setting, the running times of
the two different solvers can diverge considerably. Figure
\ref{fig:MNISTquant:right} plots the two runtimes against each other
(along with the corresponding curve length, in color). The
probabilistic solvers' slight decrease in precision is accompanied by
a decrease in computational cost by about an order of magnitude. Due
to its design, the computational cost of the probabilistic solver is
almost unaffected by the complexity of the problem (length of the
curve). Instead, harder problems simply lead to larger uncertainty. As
dimensionality increases, this advantage grows. Fig.~\ref{fig:mnist},
right, shows the running time for computations of the mean for an
increasing number of dimensions. For this experiment we use 100 images
and 5 iterations of the gradient descent scheme. As the plot shows,
the probabilistic approach is substantially faster in this experiment.

\subsection{Human Body Shape}\label{sec:bodies}

\begin{figure}
  \centering \scriptsize
  \begin{tabular}{ccc}
    $\mu - 3\sigma$ & $\mu$ & $\mu + 3\sigma$ \\
    \hspace{-4mm}
    \includegraphics[width=0.16\textwidth]{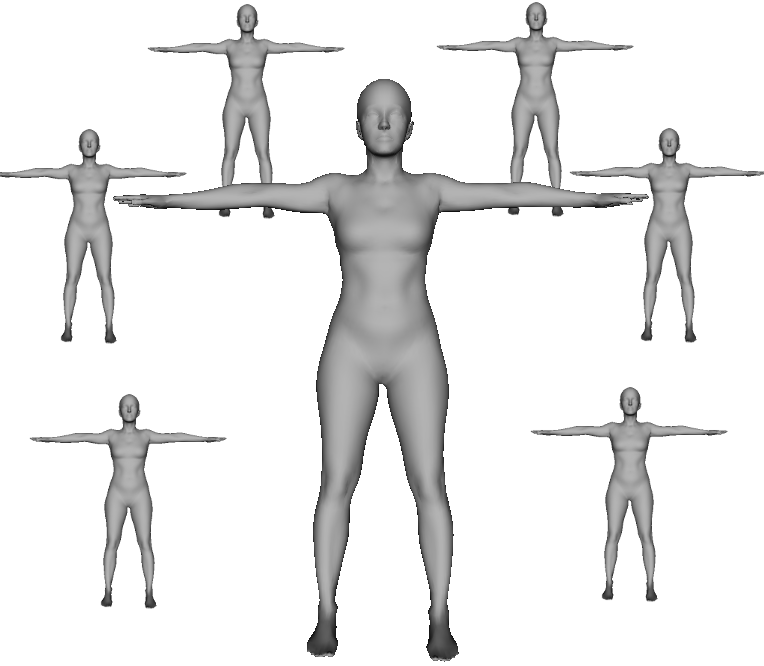} \hspace{-3mm} & \hspace{-3mm}
    \includegraphics[width=0.16\textwidth]{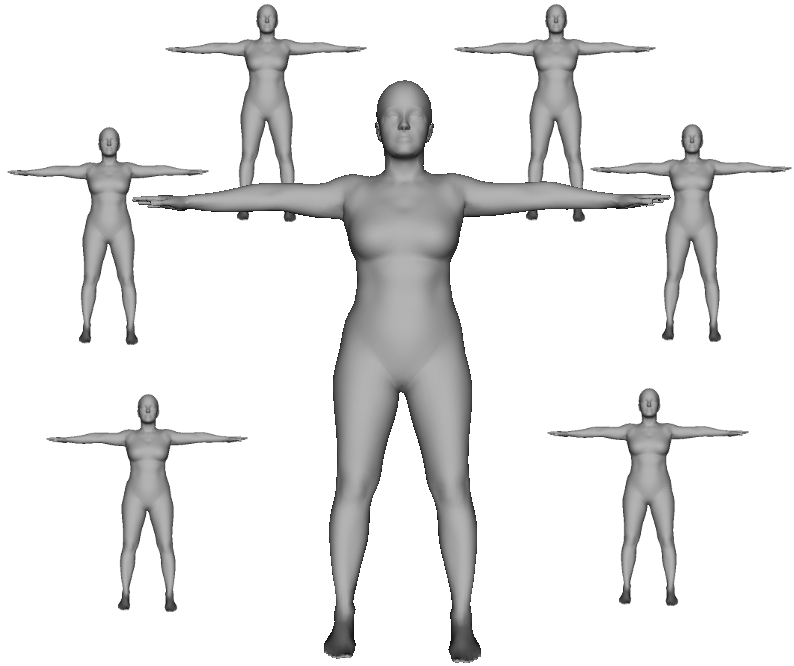}  \hspace{-3mm} & \hspace{-3mm}
    \includegraphics[width=0.16\textwidth]{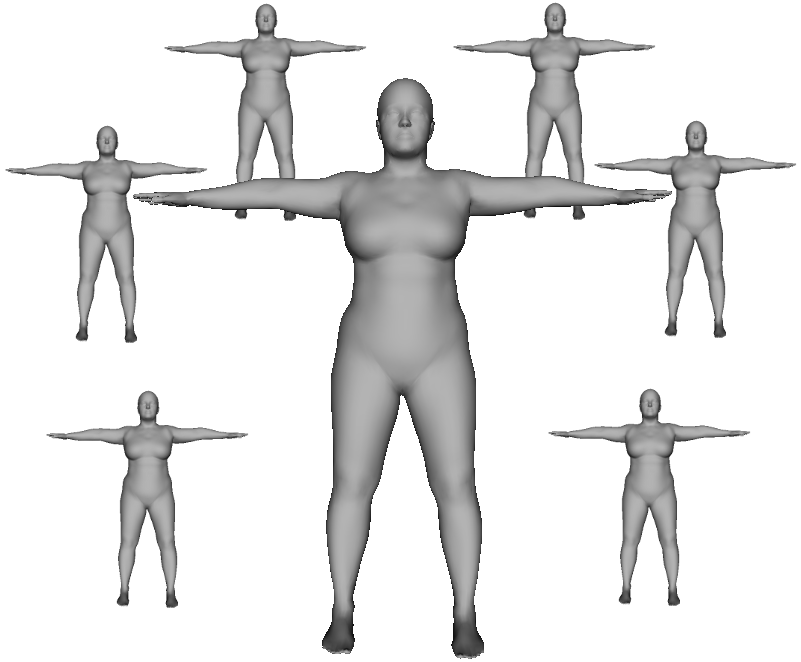}
  \end{tabular}
  \vspace{-4mm}
  \caption{Samples from the posterior over the principal geodesic
    through the body shape dataset, estimated under a metric learned
    to emphasise belly circumference. Six samples each at the
    inferred manifold mean (centre) and three standard deviations of
    the dataset in either direction (left, right).
    \emph{The variation is shown in the supplementary video}.}
  \label{fig:bodies}
\end{figure}

We also experimented with human body shape data, using $1191$ laser
scans and anthropometric measurements of the belly circumference of
female subjects \citep{caesar}. Body shape is represented using a
$50$-dimensional SCAPE model \citep{scape}, such that each scan is a
vector in $\Re^{50}$. Following \citet{hauberg:nips:2012}, we
partition the subjects in $7$ clusters by belly circumferences, learn
a diagonal \emph{Large Margin Nearest Neighbour (LMNN)} metric
\citep{weinberger09distance} for each cluster, and form a smoothly
changing metric using Eq.~\ref{eq:metric}. Under this metric 
variance is inflated locally along directions related to belly
circumference. We visualise this increased variance using PGA. Based
on the timing results above, we only consider the probabilistic
algorithm here, which achieved a runtime of $\sim$10 minutes.

Fig.~\ref{fig:bodies} shows the first principal geodesic under the
learned metric, where we show bodies at the mean as well as $\pm$
three standard deviations from the mean. Uncertainty is visualised by
six samples from the principal geodesic at each point; this uncertainty
is not described by other solvers (this is more
easily seen in the supplementary video). In general, the principal
geodesic shows an increasing belly circumference (again, see
supplementary video).


\section{Conclusion}
\label{sec:conclusions}

We have studied a probabilistic solver for ordinary differential
equations in the context of Riemannian statistics. The method can deal
with uncertain boundary and initial values, and returns a joint
posterior over the solution. The theoretical performance of such
methods is currently not as tightly bounded as that of Runge-Kutta
methods, but their structured error estimates allow a closed
probabilistic formulation of computations on Riemannian
manifolds. This includes marginalisation over the solution space,
which weakens the effect of numerical errors. This, in turn, means
computational precision, and thus cost, can be lowered. The resulting
improvement in computation time and functionality is an example for
the conceptual and applied value of probabilistically designed
numerical algorithms for computational statistics.


\subsubsection*{Acknowledgements}
S.~Hauberg is supported in part by the Villum Foundation as well as an AWS in Education Machine Learning Research Grant award from \url{Amazon.com}.

\bibliographystyle{plainnat} 
\bibliography{bibfile}

\newpage
\appendix
\onecolumn
\begin{center}
  \Large --- Supplementary Material ---
\end{center}

\section{Gaussian Process Posteriors}
\label{sec:gauss-proc-post}

Equations (10), (12), (18) and (19) in the main paper are Gaussian
process posterior distributions over the curve $c$ arising from
observations of various combinations of derivatives of $c$. These
forms arise from the following general result.\footnote{Equation A.6
  in C.E. Rasmussen and C.K.I. Williams. \emph{Gaussian Processes for
    Machine Learning.}  MIT Press, 2006} Consider a Gaussian process
prior distribution
\begin{equation}
  \label{eq:12}
  p(c) = \GP(c;\mu,k) 
\end{equation}
over the function $c$, and observations $y$ with the likelihood
\begin{equation}
  \label{eq:13}
  p(y\g c,A) = \N(y;Ac,\Lambda),
\end{equation}
with a linear operator $A$. This includes the special cases of the
selection operator $A = \delta(x-x_i)$ which selects function values
$Ac = \int \delta(x-x_i) c(x) \dd x = c(x_i)$, and the special case of
derivative operators $\de_x ^n\delta(x-x_i)$ which give $Ac=\int \de_x
\delta(x-x_i) c(x) \dd x= c^{(n)}(x_i)$. Then the posterior over any
linear map $Bc$ of the curve $c$ (including $B=\delta(x-x_j)$, giving
$Bc=c(x_j)$) is 
\begin{equation}
  \label{eq:14}
  p(Bc\g y,A) = \GP(Bc;B\mu + BkA\Trans (AkA\Trans +
  \Lambda)^{-1}(y-A\mu),BkB - BkA\Trans(AkA\Trans +
  \Lambda)^{-1}Ak B\Trans).
\end{equation}
And the marginal probability for $y$ is
\begin{equation}
  \label{eq:16}
  p(y\g A) = \int p(y\g c,A)p(c)\,dc = \N(y;A\mu,AkA\Trans + \Lambda)
\end{equation}
The classic example is that of the marginal posterior at $c(x_*)$
arising from noisy observations at $[c(x_1),\dots,c(x_N)]\Trans$. This is
the case of $B = \delta(x-x_*)$ and
$A=[\delta(x-x_1),\dots,\delta(x-x_N)]\Trans$, which gives 
\begin{align}
  \label{eq:15}
  B\mu &= \mu(x_*)\\
  A\mu &= [\mu(x_1),\dots \mu(x_N)]\Trans\\
  BkA\Trans &= \left[\iint \delta(a-x_*) k(a,b) \delta(b-x_i) \, da \,
    db\right]_{i=1,\dots,N} 
  =  [k(x_*,x_1),\dots k(x_*,x_N)]\\
  AkA\Trans &=
  \begin{pmatrix}
    k(x_1,x_1) &\cdots& k(x_1,x_N)\\
    \vdots &\ddots& \vdots\\
    k(x_N,x_1) & \cdots & k(x_N,x_N)
  \end{pmatrix}
\end{align}
and so on. All the Gaussian forms in the paper are special cases with
various combinations of $A$ and $B$.

\section{Covariance Functions}
\label{sec:covariance-functions}

The models in the paper assume a squared-exponential (aka. radial
basis function, Gaussian) covariance function between values of the
function $\f:\Re\to\Re^N$, of the form
\begin{equation}
  \label{eq:1}
  \Cov(f_i(t),f_j(t')) =
  V_{ij}\exp\left(-\frac{(t-t')^2}{2\lambda^2}\right) \ec V_{ij} k_{tt'}
\end{equation}
The calculations require the covariance between various combinations
of derivatives of the function. For clear notation, we'll use the
operator $\de\ce \de/\de t$, and the abbreviation $\delta_{tt'} \ce (t-t')/\lambda^2$
\begin{align}
  \label{eq:2}
  \Cov(f_i(t),\dot{f}_j(t')) &= V_{ij} k_{tt'} \de\Trans =
  V_{ij} \frac{t-t'}{\lambda^2}k_{tt'} = V_{ij}\delta_{tt'}k_{tt'} = - \Cov(\dot{f}_i(t),f_j(t'))\\
  \Cov(\dot{f}_i(t),\dot{f}_j(t')) &= V_{ij} \de k_{tt'}\de\Trans =
  V_{ij}\left(\frac{1}{\lambda^2} -
    \left(\frac{t-t'}{\lambda^2}\right)^2\right)k_{tt'} =
    V_{ij}\left(\frac{1}{\lambda^2}-\delta^2 _{tt'}\right)k_{tt'}\\
    \label{eq:4}
    \Cov(f_i(t),\ddot{f}_j(t')) &= V_{ij} k_{tt'} \de\Trans \de\Trans
    = V_{ij}
    \left(\left(\frac{t-t'}{\lambda^2}\right)^2-\frac{1}{\lambda^2}\right)k_{tt'}
    = V_{ij} \left( \delta^2 _{tt'} - \frac{1}{\lambda^2}\right)
    k_{tt'} = -\Cov(\dot{f}_i(t),\dot{f}_j(t'))\\
    \label{eq:5}
    \Cov(\dot{f}_i(t),\ddot{f}_j(t')) &= V_{ij}\de k_{tt'} \de\Trans
    \de\Trans = V_{ij}\left(\frac{2}{\lambda^2}\frac{t-t'}{\lambda^2}
      -\frac{t-t'}{\lambda^2}\left(\left(\frac{t-t'}{\lambda^2}\right)^2-\frac{1}{\lambda^2}\right)
    \right)k_{tt'} = V_{ij}\left(-\delta^3 _{tt'} +
      \frac{3}{\lambda^2}\delta_{tt'} \right) k_{tt'}\\
    \label{eq:6}
    \Cov(\ddot{f}_i(t),\ddot{f}_j(t')) &= V_{ij}\de\de k_{tt'}
    \de\Trans \de\Trans = V_{ij}\left(\delta^4 _{tt'}
      -\frac{6}{\lambda^2}\delta^2 _{tt'} + \frac{3}{\lambda^4}
    \right)k_{tt'}
\end{align}
Of course, all those derivatives retain the Kronecker structure of the
original kernel, because $\de(V\otimes k)=V\otimes \de k$.

\section{Inferring Hyperparameters}
\label{sec:inferr-hyperp}

Perhaps the most widely used way to learn hyperparameters for Gaussian
process models it type-II maximum likelihood estimation, also known as
evidence maximisation: The marginal probability for the observations
$y$ is $p(y\g \lambda) = \int p(y\g c) p(c\g \lambda)\d c = \N(y;\ddot
\mu_T,\de \de k_{TT}(\lambda) \de \de + \Lambda)$. Using the shorthand
$G := (\de \de k_{TT}(\lambda) \de \de + \Lambda)$, its logarithm is
\begin{align}
  \label{eq:3}
  -2\log p(y\g \lambda) = (y-\ddot \mu_T)\Trans G^{-1}(y - \ddot
  \mu_T) + \log | G | + N\log2\pi
\end{align}
To optimise this expression with respect to the length scale
$\lambda$, we use
\begin{align}
  \label{eq:7}
  -2\frac{\de\log p(y\g \lambda)}{\de \lambda^2} = - (y-\ddot
  \mu_T)\Trans G^{-1} \frac{\de G}{\de \lambda^2} G^{-1} (y-\ddot \mu_T)
  + \tr(G^{-1} \frac{\de G}{\de \lambda^2}).
\end{align}
From Equation (\ref{eq:6}), and using 
\begin{equation}
  \label{eq:9}
  \frac{\de \delta_{tt'}}{\de \lambda^2} = -\frac{\delta _{tt'}}{\lambda^2} \qqqq
  \frac{\de k_{tt'}}{\de \lambda^2} = k_{tt'} \frac{\delta^2 _{tt'}}{2}
\end{equation}
we find
\begin{align}
  \label{eq:8}
  \frac{\de G^{ij} _{tt'}}{\de \lambda^2} &= V_{ij}\left[
  \left(-\frac{4}{\lambda^2}\delta^4 _{tt'}
    +\frac{18}{\lambda^4}\delta^2 _{tt'} + \frac{6}{\lambda^6} \right)
  k_{tt'}  + \de\de k_{tt'}\de\Trans\de\Trans \frac{\delta_{tt'}
    ^2}{2}\right]\\
&=V_{ij}\left(\frac{\delta^6}{2} -\frac{7}{\lambda^2}\delta^4 _{tt'}
    +\frac{39}{2\lambda^4}\delta^2 _{tt'} + \frac{6}{\lambda^6} \right)
  k_{tt'} 
\end{align}
It is also easy to evaluate the second derivative, giving
\begin{align}
  -2\frac{\de^2\log p(y\g \lambda)}{(\de \lambda^2)^2}&=2 (y-\ddot
  \mu_T)\Trans G^{-1} \frac{\de G}{\de \lambda^2} G^{-1} \frac{\de
    G}{\de \lambda^2} G^{-1} (y-\ddot \mu_T) - \tr\left[\frac{\de
      G}{\de \lambda^2} G^{-1}\frac{\de G}{\de
      \lambda^2} G^{-1} \right]\\
  &- (y-\ddot \mu_T)\Trans G^{-1} \frac{\de^2 G}{(\de \lambda^2)^2}
  G^{-1} (y-\ddot
  \mu_T) + \tr \left[G^{-1} \frac{\de^2 G}{(\de \lambda^2)^2} \right]\\
  \text{where}\qq\frac{\de^2 G^{ij} _{tt'}}{(\de \lambda^2)^2} &=
  V_{ij}\left(-\frac{3}{\lambda^2}\delta_{tt'} ^{6} +
    \frac{35}{\lambda^4}\delta_{tt'} ^{4} -
    \frac{78}{\lambda^6}\delta_{tt'} ^{2} +\frac{18}{\lambda^8}\right)
  k_{tt'} +\frac{\delta^2}{2}\frac{\de G_{tt'} ^{ij}}{\de \lambda^2}
\end{align}
This allows constructing a Newton-Raphson optimisation scheme for the
length scale of the algorithm.

\end{document}